\documentclass[11pt]{article}
\usepackage[preprint]{acl}
\usepackage{times}
\usepackage{latexsym}
\usepackage[T1]{fontenc}
\usepackage[utf8]{inputenc}

\usepackage{amsmath,amsfonts,bm}

\def\eqref#1{equation~\ref{#1}}
\def\1{\bm{1}}

\DeclareMathAlphabet{\mathsfit}{\encodingdefault}{\sfdefault}{m}{sl}
\SetMathAlphabet{\mathsfit}{bold}{\encodingdefault}{\sfdefault}{bx}{n}

\DeclareMathOperator*{\argmax}{arg\,max}

\usepackage{hyperref}
\usepackage{url}
\usepackage{subcaption}
\usepackage{amsmath}
\usepackage{algorithm}
\usepackage{algpseudocode}
\usepackage{booktabs} 
\usepackage{tcolorbox}
\usepackage{wrapfig}
\usepackage{graphicx}
\usepackage{tabularx}
\usepackage{array}
\usepackage{xcolor}
\usepackage{caption}

\title{Evaluating Reward Model Generalization via Pairwise Maximum Discrepancy Competitions}

\author{
\textbf{Shunyang Luo}$^{1*}$,
\textbf{Peibei Cao}$^{2}$\thanks{Equal contribution.},
\textbf{Zhihui Zhu}$^{3}$,
\textbf{Kehua Feng}$^{3}$,
\textbf{Zhihua Wang}$^{4}$,
\textbf{Keyan Ding}$^{3}$\thanks{Corresponding authors.} \\
$^{1}$ZJU-UIUC Institute, Zhejiang University \\
$^{2}$School of Artificial Intelligence, Nanjing University of Information Science and Technology \\
$^{3}$ZJU-Hangzhou Global Scientific and Technological Innovation Center, Zhejiang University \\
$^{4}$City University of Hong Kong \\[0.05em]
\texttt{shunyang.22@intl.zju.edu.cn, dingkeyan@zju.edu.cn}
}

\begin{document}
\raggedbottom
\maketitle

\begin{abstract}
Reward models (RMs) are central to aligning large language models, yet their practical effectiveness hinges on \emph{generalization} to unseen prompts and shifting distributions. Most existing RM evaluations rely on static, pre-annotated preference datasets, which provide limited coverage and often fail to faithfully assess generalization in open-world settings. 
We introduce {Pairwise Maximum Discrepancy Competition (PMDC)}, a dynamic and annotation-efficient framework for evaluating RM generalization using a large, unlabeled, open-domain prompt pool. PMDC \textit{actively} selects prompt--response pairs that maximize disagreement between two RMs, yielding a compact set of highly contentious test cases. These cases are adjudicated by an oracle, and the resulting outcomes are aggregated via a Bradley--Terry model to produce a global ranking and pairwise win-rate landscape of RMs.
We apply PMDC to re-evaluate $10$ representative RMs and observe substantial rank reshuffling compared with conventional benchmarks. Qualitative analyses further uncover systematic generalization failures, providing valuable insights for improving reward modeling.
\end{abstract}

\section{Introduction}
Reward models (RMs) are a cornerstone of modern alignment pipelines, enabling large language models (LLMs) to internalize complex human preferences through reinforcement learning from human feedback (RLHF) \citep{christiano2023, stiennon2022}. By learning to predict human preferences between pairs of model responses, RMs provide scalable training signals that guide LLMs toward desirable behaviors across various domains, including instruction following, reasoning, and safety \citep{bai2022training,ouyang2022training,feng2025sampleefficienthumanevaluationlarge}. Their real-world effectiveness, however, is determined by their ability to generalize to unseen prompts and shifting data distributions. 

However, prevailing evaluation benchmarks \citep{lambert2024rewardbench} for RMs predominantly rely on static, pre-annotated datasets offering limited coverage and are inadequate for faithfully assessing generalization in open-world settings. 
These conventional evaluation resources suffer from several critical limitations. First, their restricted coverage of the potential prompt and behavioral space impedes the assessment of model generalization to novel domains or edge-case scenarios. 
Second, the human annotations underpinning these datasets are typically sourced from specific demographic groups or constrained task contexts, potentially introducing biases that do not accurately reflect broader human judgment. Third, the fixed and publicly accessible nature of these test sets introduces inherent risks of overfitting, both explicit and implicit, 
where models may be optimized for benchmark performance without achieving meaningful improvements in alignment quality or robustness \citep{gao2022scaling,zhong2025comprehensive,kim2025rethinking}.

To address these challenges, we propose the \textit{Pairwise Maximum Discrepancy Competition} (PMDC), a dynamic and cost-efficient framework for evaluating generalization of RMs. Inspired by discrepancy-driven evaluation paradigms in computer vision \citep{saito2018maximum}, PMDC actively identifies prompt-response pairs that maximize disagreement between two RMs from a large, unlabeled, open-domain prompt pool, producing a focused set of highly contentious test cases. Such contentious instances are adjudicated by an oracle, here, a powerful LLM serving as a scalable proxy for human judgment.  
The results are aggregated via a Bradley-Terry (BT)  model \citep{bradley1952rank} to generate a global ranking and detailed win-rate landscape.

Crucially, PMDC shifts the evaluation paradigm from static benchmarking to active and adaptive probing. 
This approach offers two key advantages. First, it enables \textit{dynamic evaluation} by adaptively sampling test cases from an open-domain prompt pool and utilizing responses generated by a diverse set of LLMs, thereby facilitating the detection of out-of-distribution failures and enhancing generalization assessment. Second, it ensures \textit{annotation efficiency} by submitting only the most discriminative sample pairs to the oracle for judgment, which significantly reduces annotation costs. We apply PMDC to re-evaluate $10$ representative RMs, observing substantial rank reshuffling compared to conventional benchmarks and uncovering systematic generalization failures that provide actionable insights for improving reward modeling.

The main contributions of this work are threefold:

\begin{itemize}
\item \textbf{The PMDC Framework:} A novel paradigm that moves beyond static benchmarks to dynamically evaluate RM generalization through active, discrepancy-driven sampling.

\item \textbf{An Actively Probed Dataset:} A compact, high-quality evaluation set, generated by identifying points of maximum inter-model disagreement, which facilitates discrimination-rich assessment and can enhance downstream alignment. 

\item \textbf{New Empirical Insights:} A re-evaluation of state-of-the-art RMs that reveals significant ranking inconsistencies with prior benchmarks and provides valuable diagnostics for generalization failures.
\end{itemize}

\section{Related Works}

\subsection{Reward Model Benchmarks}
Initial efforts to evaluate RMs established the paradigm of measuring preference prediction accuracy on curated, static datasets. Benchmarks like RewardBench provide foundational, closed-set assessments of RM capabilities \citep{lambert2024rewardbench}. However, subsequent research has questioned whether accuracy on such narrow datasets reliably correlates with downstream alignment performance or generalization~\citep{wen2025rethinkingrewardmodelevaluation, levine2024baselineanalysisrewardmodels}. In response, more comprehensive benchmarks have emerged, including RM-Bench \citep{liu2024rmbenchbenchmarkingrewardmodels} and RewardBench 2 ~\citep{malik2025rewardbench2advancingreward}, which assess RMs on nuanced capabilities like discerning subtlety and resisting stylistic bias. 
This paradigm has further extended into specialized domains, reflecting the growing application scope of reward modeling. 
Recent benchmarks now assess multilingual~\citep{gureja2025mrewardbenchevaluatingrewardmodels}, vision-language~\citep{yasunaga2025multimodalrewardbenchholisticevaluation, li2025vlrewardbenchchallengingbenchmarkvisionlanguage}, and embodied agent~\citep{men2025agentrewardbenchunifiedbenchmarkreward} RMs.
Concurrently, the adoption of powerful LLMs as reward models or preference judges has gained significant traction~\citep{zheng2023judging,dong2024can}. To formalize and standardize the assessment of these LLM-based evaluators, several dedicated benchmarks have been proposed~\citep{thakur2025judgingjudgesevaluatingalignment,murugadoss2024evaluatingevaluatormeasuringllms,tan2025judgebenchbenchmarkevaluatingllmbased,zhou2025evaluatingjudgesevaluatorsjetts}. These frameworks evaluate critical dimensions such as alignment with human preferences, robustness to varying instruction complexities, and consistency across diverse evaluation scenarios, establishing much-needed rigor in judge-style model assessment.

\subsection{Maximum Discrepancy Competition} 
Beyond static benchmarks, it is crucial to actively and efficiently probe for model weaknesses. The Maximum Discrepancy (MAD) competition framework provides a powerful methodology for this task \citep{ma2020gmad}. Instead of relying on a fixed test set, MAD adaptively samples data points that cause the largest disagreement between two or more competing models. This principle has been successfully applied to expose failures and compare models in diverse domains, including objective image quality \citep{ma2016cvpr} and semantic segmentation \citep{yan2021exposingsemanticsegmentationfailures}. More recently, this sample-efficient approach has been adapted for the human evaluation of large language models, demonstrating its effectiveness in identifying the most informative examples to distinguish between high-performing models~\citep{feng2025sampleefficienthumanevaluationlarge}. Our work is inspired by this adversarial, comparative approach to develop a more robust and efficient evaluation protocol for reward models.

\begin{figure*}[t]
    \centering
    \includegraphics[width=\textwidth]{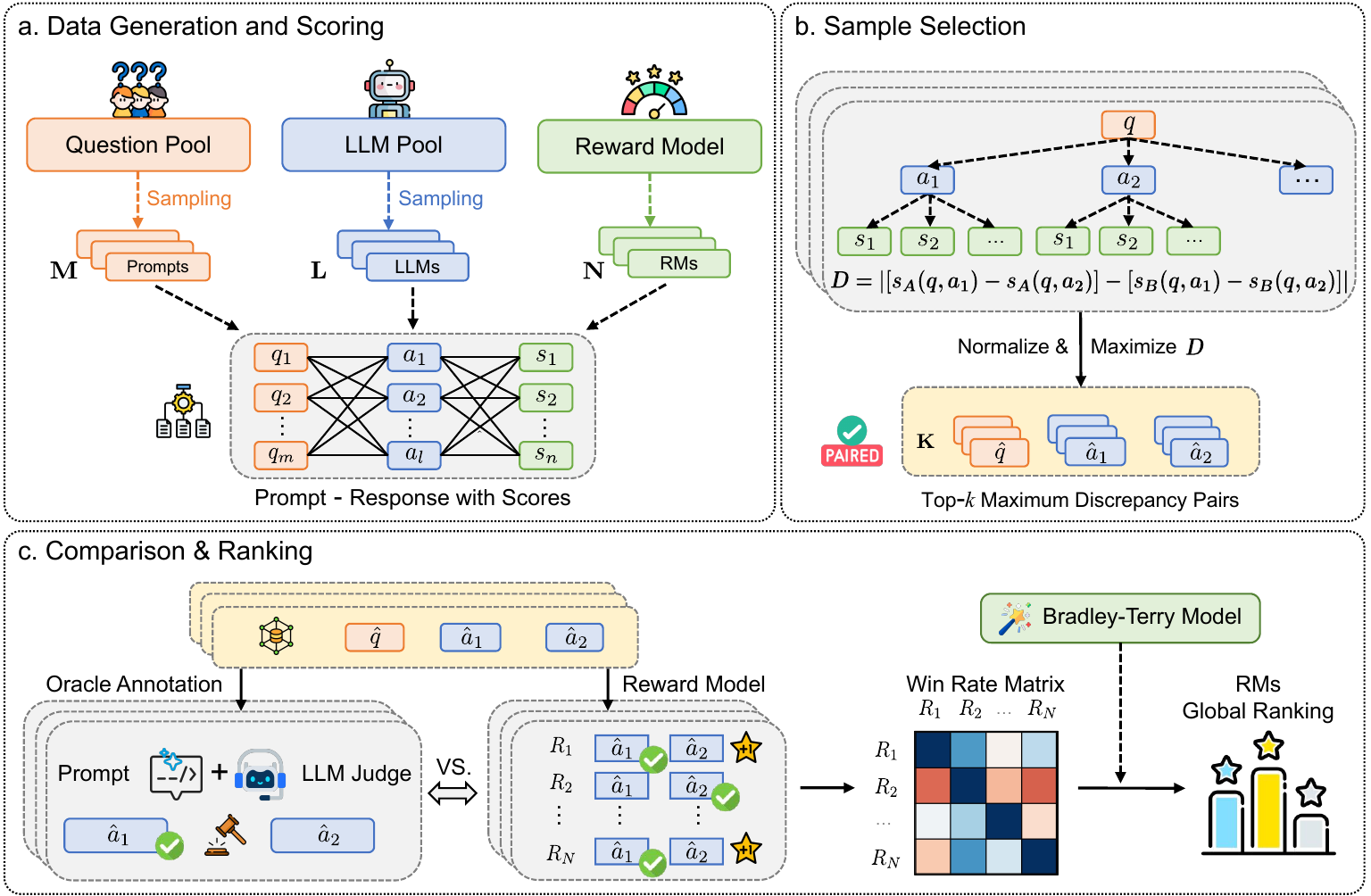}
    \caption{Overview of the proposed PMDC framework. 
    (a)~\textit{Data generation and scoring}: Sample prompts and LLMs to build prompt-response pairs, which are then scored by reward models (RMs).
    (b)~\textit{Sample selection}: Based on the Maximum Discrepancy Competition principle, select top-$k$ pairs (with maximum RM preference discrepancy) to form an evaluation subset.
    (c)~\textit{Comparison \& ranking}: Annotate the selected QA pairs with an Oracle (i.e., LLM-based Judge) to rank responses, compare Oracle results with RMs to build a win-rate matrix, and convert the pairwise comparisons into RMs’ global ranking using the Bradley-Terry model.}
    \label{fig:pmdc_framework}
\end{figure*}

\section{Proposed PMDC}
This section details the PMDC, our dynamic framework for evaluating RM generalization. 
As illustrated in Figure~\ref{fig:pmdc_framework}, our method proceeds in three core stages: 1) constructing a diverse, unlabeled evaluation pool; 2) actively selecting the most informative, high-discrepancy test cases; and 3) adjudicating these cases and aggregating the results into a global ranking.
% dynamically samples prompts from diverse datasets, generates responses using multiple LLMs, identifies maximum discrepancy pairs, obtains oracle judgments for contentious cases, and finally produces global rankings.

\subsection{Data Generation and Scoring }
We begin by constructing a diverse evaluation corpus $\mathcal{X}$. We sample $M$ prompts from a comprehensive, open-domain pool to ensure broad topical and stylistic coverage. Concurrently, we select $L$ distinct LLMs from a diverse model pool. For each prompt $q_j$, we generate $L$ candidate responses using the selected LLMs, yielding a dataset $\mathcal{X} = \{(q_j, \{a_j^{(m)}\}_{m=1}^L)\}_{j=1}^M$ that captures a wide range of response strategies.

% The data generation begins with the construction of a diverse evaluation corpus through systematic sampling from multiple data sources. We first sample $M$ prompts from a comprehensive prompt pool comprising prompts aggregated from established benchmarks to ensure broad topical coverage. Concurrently, we sample $L$ LLM from a diverse model pool.
% For each sampled prompt, we generate candidate responses using the selected LLMs, resulting in $M$ prompts each with $L$ potential responses. This process yields a dataset of prompt-response pairs that captures diverse response strategies and styles. The resulting collection forms our evaluation dataset $\mathcal{X} = \{(q_j, \{a_j^{(m)}\}_{m=1}^L)\}_{j=1}^M$.

Each prompt-response pair $(q, a)$ is then evaluated by $N$ distinct RMs $\mathcal{R} = \{R_i\}_{i=1}^N$. Each RM $R_i$ assigns a real-valued score $s_i(q,a)$ reflecting its assessment of the response's quality. To enable a fair comparison across RMs with potentially different output scales, we apply min-max normalization per model across the entire dataset:
\begin{equation}
s'_{i}(q, a) = \frac{s_i(q, a) - \min_i}{\max_i - \min_i},
\end{equation}
where $\min_i$ and $\max_i$ represent the minimum and maximum scores produced by $R_i$ over the dataset, respectively.
These normalized scores are converted into discrete preferences:
\begin{equation}
\text{Pref}(R_i; q, a_1, a_2)=
\begin{cases}
1 & \text{if } s'_i(q,a_1) > s'_i(q,a_2),\\
0 & \text{otherwise}.\\
\end{cases}
\end{equation}

\subsection{Sample Selection}
The objective is to efficiently evaluate and rank $N$ RMs on the dataset $\mathcal{X}$. Conventional evaluation methods heavily rely on static and pre-annotated datasets, which frequently fail to measure the generalization ability of RMs when confronted with unseen prompts in open-world settings. The standard process for evaluating RMs on new datasets consists of three stages. First, a small dataset $\mathcal{S}$ must be pre-selected. Second, predictions are generated by processing $\mathcal{S}$ through the RMs. Third, human evaluation is performed on these outputs to compare relative model performance. The RM that achieves the highest average subjective rating across $\mathcal{S}$ is considered superior.
However, this evaluation paradigm is labor-intensive, expensive, and challenging to scale, posing significant practical limitations for efficient assessment of reward models.

% \noindent\textbf{Maximum Discrepancy Sample Selection.}
Following the MAD principle \citep{wang2008maximum}, we aim to evaluate the RM generalization by adaptively selecting a minimal yet highly informative subset of prompt-response pairs.
We begin by considering the \textit{simplest} scenario, where two RMs $R_A$ and $R_B$ are being compared under an oracle budget that permits the judgment of only one prompt and its corresponding response pair $(q,\{a_1, a_2\})\in \mathcal{X}$. The core challenge thus reduces to: How can we automatically select the most informative sample from a large pool of candidates such that the relative performance between $R_A$ and $R_B$ can be most effectively discerned?

According to the MAD competition methodology, PMDC selects the prompt-response pair $(\hat{q},\{\hat{a}_1, \hat{a}_2\}) \in \mathcal{X}$ that best differentiates between RMs $R_A$ and $R_B$:
\begin{equation}
\label{eq:discrepancy}
\resizebox{\linewidth}{!}{$
\begin{aligned}
(\hat{q},\{\hat{a}_1, \hat{a}_2\})
&= \argmax_{(q,\{a_1, a_2\}) \in \mathcal{X}} \big|
    [s'_A(q,a_1)-s_A'(q,a_2)]
\\
&\qquad
    -[s_B'(q,a_1)-s_B'(q,a_2)]
\big|.
\end{aligned}
$}
\end{equation}
where $s'_A(q,a_1)-s'_A(q,a_2)$ represents the preference score difference assigned by model $R_A$ to the response pair $\{a_1, a_2\}$, with a larger positive value indicating a stronger preference for $a_1$ over $a_2$. The same applies to $s_B'(q,a_1)-s_B'(q,a_2)$ for $R_B$.

Then, we extend this idea to compare $R_A$ and $R_B$ over a small subset $\mathcal{S} \subset \mathcal{X}$ comprising $K$ prompt–response pairs with the highest discrepancy values, as computed by  Eq.~\ref{eq:discrepancy}. The $k$-th pair is selected iteratively using:
\begin{equation}
\label{eq:discrepancy2}
\resizebox{\linewidth}{!}{$
\begin{aligned}
(\hat{q},\{\hat{a}_1, \hat{a}_2\})^{(k)}
&= \arg\max_{(q,\{a_1, a_2\}) \in \mathcal{X}\setminus\mathcal{S}}
\Big|
    [s'_A(q,a_1)-s'_A(q,a_2)]
\\
&\qquad
    - [s'_B(q,a_1)-s'_B(q,a_2)]
\Big|.
\end{aligned}
$}
\end{equation}
where $\mathcal{S}=\{(\hat{q},\{\hat{a}_1, \hat{a}_2\})\}_{i=1}^{k-1}$ contains the previously chosen $k-1$ pairs. Each newly selected pair is incorporated into $\mathcal{S}$ for subsequent iterations.

\subsection{Comparison \& Ranking}
The oracle assessment of the preferences from $R_A$ and $R_B$ for a given pair $(q,\{a_1, a_2\})$ leads two plausible results:
\begin{itemize}
 \item The oracle’s judgment is consistent with that of $R_A$ (or $R_B$). In this case, PMDC successfully identifies the most informative prompt–response pair for discriminating between the two models, thereby enabling a conclusive performance ranking.

 \item The oracle cannot determine a superior response, which is possible in open-world scenarios. 
 Although the selected prompt-response pair $(\hat{q},\{\hat{a}_1, \hat{a}_2\})$ may reveal divergent strengths (or weaknesses) of $R_A$ and $R_B$, but contributes less to their relative performance ranking.
\end{itemize}

Given $N$ RMs, PMDC chooses top-$k$ prompt-response pairs for each of the $\binom{N}{2}$ model pairs,  resulting in a final evaluation set $\mathcal{D}$ of size $N(N-1)K/2$. Notably, the size of $\mathcal{D}$ is independent of the size of the input domain $\mathcal{X}$, allowing PMDC to benefit from an expanded $\mathcal{X}$ with broader prompt–response coverage.

For the Oracle assessment, PMDC employs a two-alternative forced choice (2AFC) paradigm. Each prompt–response pair
$(q,\{a_1, a_2\}) \in \mathcal{S}$ is presented to the oracle alongside the outputs of two competing RMs, $R_A$ and $R_B$. The oracle is required to select the preferred response.
The collected judgments are compiled into an  $N\times N$ win-count matrix $W$, where $W_{i,j}$ records the number of votes for $R_i$ and against $R_j$.
The symmetrized win rate matrix is computed as:
\begin{equation}
\label{eq:winrate}
P_{i,j} = \frac{W_{i,j}}{W_{i,j}+W_{j,i}+\varepsilon}, \quad P_{i,i}=0.5,
\end{equation}
where $\varepsilon$ is a small smoothing constant, ensuring $P_{i,j}+P_{j,i}\approx 1$ off-diagonal and neutral diagonal.

% \noindent\textbf{Global Ranking}
We employ the BT model to infer the global ranking of $\mathcal{R}$. Specifically, we let $\xi$ be the vector of global ranking scores $[\xi_1, \ldots, \xi_n]$, and define the probability of $R_i$ being preferred over $R_j$ as
\begin{equation}
P_{i,j} = \frac{1}{1 + \exp(\xi_j - \xi_i)}.
\end{equation}

We estimate the global scores by maximizing regularized log-likelihood with Broyden-Fletcher-Goldfarb-Shanno (BFGS) \citep{hunter2004mm}, and applying L2 penalty ($\lambda=10^{-6}$) for numerical stability and fixing $\xi_1=0$ for identifiability:

\begin{equation}
\label{eq:bt_likelihood}
\resizebox{\linewidth}{!}{$
\begin{aligned}
\log \mathcal{L}(\boldsymbol{\xi})
&= \sum_{(i,j) \in \mathcal{C}}
\Big[
    w_{ij} \log P_{i,j}
    + w_{ji} \log P_{j,i}
\Big]
- \lambda \sum_{k=2}^{n} \xi_k^2 .
\end{aligned}
$}
\end{equation}

We summarize the proposed PMDC in Algorithm~\ref{alg:pmdc} (See Appendix).

\begin{table*}[t]
\centering
\setlength{\tabcolsep}{7pt}
\begin{tabular}{lccc}
\hline
Model & Rank & BT ranking score & Agreement (\%) \\
\hline
Skywork-Reward-Gemma-2-27B & 1 & 2.488 & 91.6 \\
QRM-Gemma-2-27B & 2 & 1.271 & 74.8 \\
Reward-Model-Mistral-7B-instruct-unified & 3 & 0.753 & 65.3 \\
URM-LLaMa-3.1-8B & 4 & 0.065 & 49.9 \\
ArmoRM-Llama3-8B-v0.1 & 5 & 0.000 & 48.3 \\
Skywork-Reward-Llama-3.1-8B & 6 & -0.185 & 44.1 \\
Skywork-Reward-V2-Qwen3-8B & 7 & -0.455 & 37.6 \\
Reward-Model-Deberta-v3-large-v2 & 8 & -0.507 & 36.7 \\
Skywork-Reward-V2-Llama-3.2-3B & 9 & -1.007 & 26.0 \\
Skywork-Reward-V2-Llama-3.1-8B & 10 & -1.059 & 24.9 \\
\hline
\end{tabular}
\caption{Global ranking results. Higher agreement indicates better oracle consistency on contentious Maximum Discrepancy samples.}
\label{tab:global_ranking_summary}
\end{table*}

\section{Experiments}
We conduct a comprehensive evaluation of the PMDC framework to assess its capability for dynamically evaluating RM generalization. 
Section~\ref{sec:experimental} details our setup, including datasets, reward models, and evaluation metrics. Section~\ref{sec:results} presents PMDC’s global rankings and compares them with established benchmarks. Section~\ref{sec:analysis} analyzes PMDC’s sensitivity to key design choices, such as top-$k$ selection, oracle judge, and sampling randomness. 
Section~\ref{sec:case_studies} presents qualitative case studies of PMDC-selected comparisons.
Finally, Section~\ref{sec:improving} demonstrates how PMDC-identified samples can improve reward models via targeted fine-tuning.

\subsection{Experimental Setup} \label{sec:experimental}

\paragraph{Dataset} We first compile a large, unlabeled prompt pool by aggregating prompts from six LLM benchmarks to ensure broad topical coverage: 1) MMLU (Massive Multitask Language Understanding)~\citep{hendryckstest2021}, 2) GSM8K (Grade School Math)~\citep{cobbe2021gsm8k}, 3) HumanEval~\citep{chen2021evaluating}, 4) AlpacaEval~\citep{alpaca_eval}, 5) TruthfulQA~\citep{lin2022truthfulqameasuringmodelsmimic}, 6) HellaSwag~\citep{zellers2019hellaswag}.
To mitigate potential data contamination (e.g., prompts appearing in RM pretraining or preference data), we rewrite prompts via instruction evolution \citep{zeng2024automatic}, which progressively transforms prompts into more diverse and higher-complexity variants. 
We further construct a pool of $20$ state-of-the-art LLMs spanning multiple model families and providers. 
For each prompt, we randomly sample several models from this pool to generate candidate responses, enabling flexible composition of prompt--response pairs.

\paragraph{Reward Models} We evaluate $10$ representative RMs, covering different architectures, training paradigms, and parameter scales: 1) Skywork-Reward-Gemma-2-27B~\citep{liu2024skywork}, 2) QRM-Gemma-2-27B~\citep{dorka2024quantile}, 3) Reward-Model-Mistral-7B-instruct-unified~\citep{yang2024regularizing}, 4) URM-LLaMa-3.1-8B~\citep{lou2025uncertainty}, 5) ArmoRM-Llama3-8B-v0.1~\citep{ArmoRM}, 6) Skywork-Reward-Llama-3.1-8B~\citep{liu2024skywork}, 7) Skywork-Reward-V2-Qwen3-8B~\citep{liu2025skywork}, 8) Reward-Model-Deberta-v3-large-v2~\citep{OpenAssistant_reward-model-deberta-v3-large-v2}, 9) Skywork-Reward-V2-Llama-3.2-3B~\citep{liu2025skywork}, 10) Skywork-Reward-V2-Llama-3.1-8B~\citep{liu2025skywork}. Among them, models 1) and 2) are large-scale 27B parameter models that demonstrate superior performance, while models 3)-6) represent mid-scale instruction-tuned variants, and models 7)-10) are recent Skywork-V2 series models with various base architectures.

\paragraph{Oracle} We employ Claude-Sonnet-4 as our oracle judge via the API endpoint. The oracle uses a systematic prompt design with structured instructions to ensure consistent and reliable judgments. The full prompt can be found in Appendix~\ref{app:oracle_prompt}.

\paragraph{Evaluation Metrics}
In addition to the global ranking from the BT model, we also report the oracle agreement rate, which measures the proportion of Maximum Discrepancy samples where an RM's preference aligns with the oracle's judgment:
\begin{equation}
\label{eq:oracle_agreement}
\resizebox{\linewidth}{!}{$
\begin{aligned}
\text{Agreement}(R_i)
&= \frac{1}{|\mathcal{S}_i|}
   \sum_{\mathbf{s} \in \mathcal{S}_i}
   \mathbb{I}\big[
       \text{Pref}(R_i,\mathbf{s})
       = \text{Oracle}(\mathbf{s})
   \big].
\end{aligned}
$}
\end{equation}
where $\mathcal{S}_i$ is the set of maximum discrepancy samples involving RM $R_i$.
A higher value indicates that the RM is more reliable when evaluating challenging samples characterized by high inter-model disagreement.

\paragraph{Implementation Details}
For each experiment, we randomly select $1,000$ prompts from the compiled prompt pool. For each selected prompt, we generate $5$ candidate responses by randomly sampling five LLMs from the LLM pool.
For each reward model pair, we systematically select the top-$k$ QA pairs with the highest reward score discrepancy across all $1,000 \times 5 = 5,000$ QA pairs using our normalized score difference metric (Eq. ~\ref{eq:discrepancy}), where $k$ is a configurable hyperparameter (default $k=10$). With $10$ RMs, this yields $\binom{10}{2} \times k = 45 \times 10 = 450$ Maximum Discrepancy samples in total. 

\begin{figure}[t]
    \centering
    \includegraphics[width=0.48\textwidth]{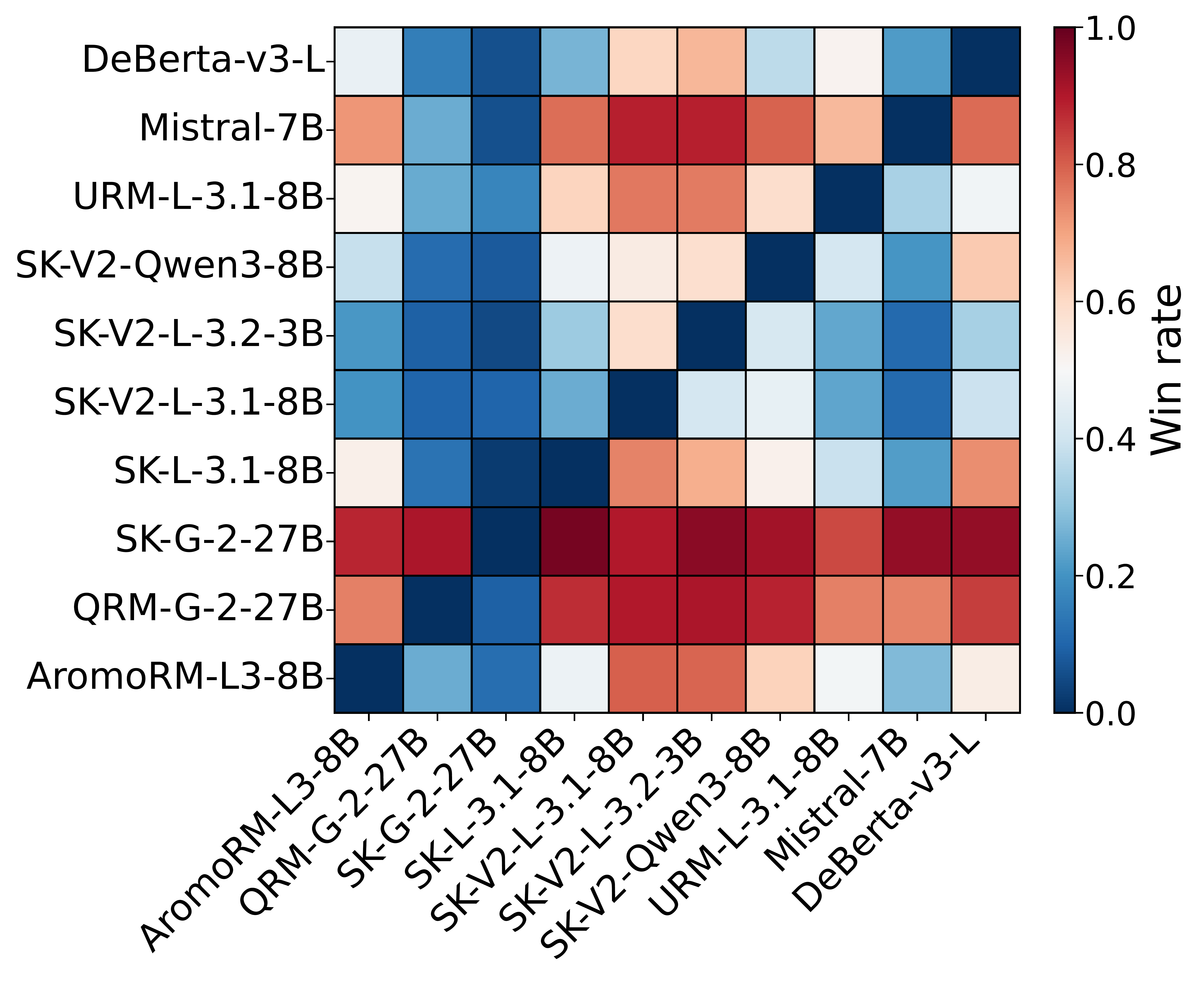}
    \caption{Pairwise win-rate heatmap across RMs on Maximum Discrepancy samples.}
    \label{fig:win_rate_heatmap}
\end{figure}

\subsection{Main Results} \label{sec:results}

\paragraph{Global Ranking Results}
Table~\ref{tab:global_ranking_summary} presents the global ranking of the evaluated models, reporting both their BT ranking score and oracle agreement rates. These results provide a direct assessment of generalization by ranking models based on their performance on actively selected, contentious test cases. 
As expected, higher BT scores correspond to greater consistency with the oracle's judgments, confirming the reliability of the estimated preferences. The pairwise win-rate landscape in Figure~\ref{fig:win_rate_heatmap} provides a more detailed view of model performance.
The results reveal a clear performance hierarchy with Skywork-Reward-Gemma-2-27B emerging as the top performer, achieving an oracle agreement rate of $91.6\%$ and the highest BT ranking score of $2.488$. This indicates consistently strong alignment with human-like judgments across contentious evaluation scenarios. QRM-Gemma-2-27B and Reward-Model-Mistral-7B-instruct-unified follow in second and third place, respectively. The dominance of Gemma-2-27B-based models in the top rankings suggests that scale and architecture significantly influence reward modeling capability.
Oracle agreement rates complement the BT ranking by measuring absolute reliability on contentious samples, with the strong positive correlation confirming the robustness of our evaluation framework.

\begin{figure}[t]
    \centering
    \includegraphics[width=\linewidth]{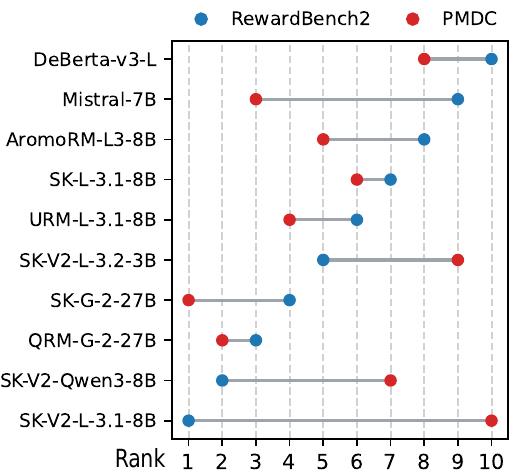}
    \caption{Rank comparison between PMDC and RewardBench2. Horizontal lines connect each reward model’s ranking under RewardBench2 (blue) and PMDC (red).}
    \label{fig:bench}
\end{figure}
    
\paragraph{Comparison Against Established Benchmarks}
To assess whether PMDC provides a more faithful evaluation of generalization in open-world settings, we compare its rankings against those from RewardBench2, a challenging held-out evaluation track. This comparison, illustrated in Figure~\ref{fig:bench}, reveals substantial rank reshuffling. While broad trends are consistent for some models, significant discrepancies emerge. A striking example is Skywork-Reward-V2-Llama-3.1-8B, which performs markedly worse under PMDC's evaluation, indicating that static benchmarks may overestimate its robustness when confronted with the novel, contentious scenarios that PMDC actively seeks out (see Appendix~\ref{app:case_studies}).

This divergence validates PMDC's core methodological premise. Unlike static benchmarks that assess recall on a fixed data distribution, PMDC actively probes for contentious edge cases. By evaluating generalization under distribution shift, it reduces sensitivity to benchmark overfitting and offers a complementary, diagnostics-rich perspective on RM robustness that conventional evaluations often miss.

\subsection{Model and Result Analysis}\label{sec:analysis}

\begin{figure}[t]
    \centering
    \includegraphics[width=\linewidth]{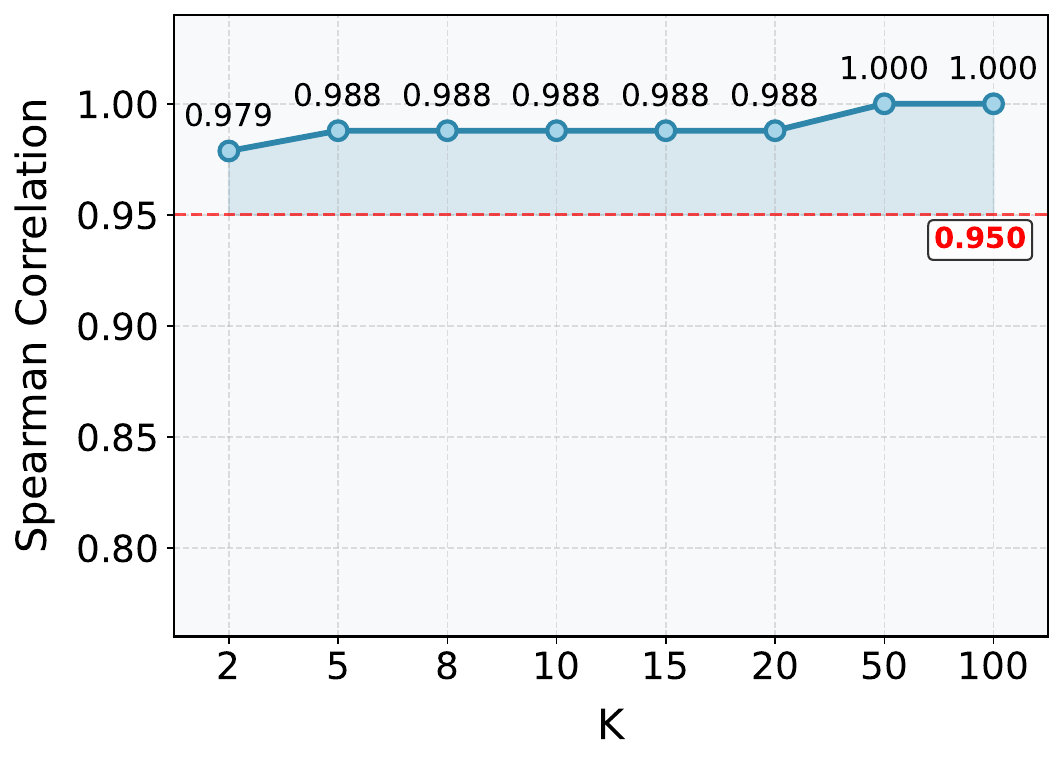}
    \caption{Spearman correlation of PMDC ranks across top-$k$ values. The dashed red line at 0.95 highlights high rank consistency.}
    \label{fig:spearman_correlation}
\end{figure}

\paragraph{Sensitivity of Top-$k$} 
To evaluate ranking stability, we vary the number of selected high-discrepancy pairs (top-$k$) from $2$ to $100$, using the ranking at $k=100$ as the reference to compute Spearman correlation for each $k$. The results show that even with $k=5$, the obtained rankings are highly consistent with those using $k=100$ (see Figure~\ref{fig:spearman_correlation}). While very low $k$ values increase variance, they can help surface rare edge-case disagreements. In practice, a moderate $k \geq 5$ provides an optimal balance, ensuring both the annotation efficiency and the ranking robustness central to the PMDC framework.

\paragraph{Annotation Efficiency Analysis} 
The PMDC framework achieves remarkable annotation reduction compared to exhaustive pairwise evaluation. In our experiments, each of the $1,000$ questions contains $5$ candidate responses, requiring $\binom{5}{2} = 10$ pairwise comparisons per question under traditional evaluation, resulting in $1,000 \times 10 =10,000$ total annotations. In contrast, PMDC with $k=10$ only requires $\binom{10}{2} \times 10=450$ comparisons, a $95.5\%$ reduction in annotation cost.
This dramatic efficiency gain demonstrates PMDC’s practical utility for large-scale reward model evaluation while preserving ranking fidelity.

\paragraph{Sensitivity of LLM Judge} 
To assess the robustness of PMDC against potential biases introduced by the choice of oracle, we evaluated the same set of Maximum Discrepancy samples (200 pairs in total) using three distinct LLM judges: Claude-Sonnet-4, Gemini-2.5-Pro, and GLM-4-Plus. In addition, we established a human oracle by asking three NLP researchers to independently evaluate each pair and aggregating their decisions via majority vote. 
Across LLM judges, the judgements are highly consistent, with agreement rates of $95.0\%$ (Claude vs. Gemini), $96.0\%$ (Claude vs. GLM), and $94.5\%$ (Gemini vs. GLM). Importantly, all LLM judges also exhibit strong agreement with the human oracle, achieving agreement rates of $94.5\%$ (Claude), $92.5\%$ (Gemini), and $93.0\%$ (GLM).
These results suggest that PMDC’s conclusions are largely insensitive to the specific oracle employed, and that LLM-based judges provide preference signals that closely align with human judgments.

\begin{figure}[t]
    \centering
    \includegraphics[width=\linewidth]{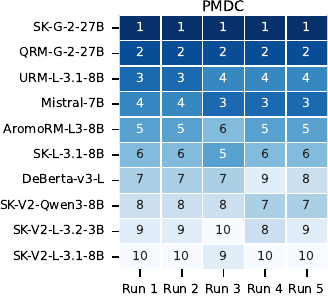}
    \caption{PMDC's rank across five independent runs. The heatmap shows the rank of each RM in each run, with rank values annotated in individual cells.}
    \label{fig:consistency}
\end{figure}

\paragraph{Result Consistency Analysis}
To assess the robustness of PMDC, we conducted five independent evaluation runs using different random seeds for prompt and LLM sampling. As illustrated in Figure~\ref{fig:consistency}, the global rankings remain highly stable across all runs, with most models maintaining identical or adjacent positions. In contrast, when using random sampling, i.e., selecting $k=10$ pairs per model pair randomly from the same pool, the resulting rankings exhibit significantly higher variance across runs, as shown in Figure~\ref{fig:random_consistency}. This instability arises because random samples often fail to capture meaningful points of disagreement between reward models, leading to noisy and inconsistent comparisons. Instead, our PMDC approach produces reliable and reproducible evaluations.

\subsection{Case Studies}
\label{sec:case_studies}
The samples identified by PMDC naturally expose systematic evaluation divergences and contrasting assessment patterns across reward models. We present five representative cases from our empirical evaluation data that reveal how different reward models exhibit distinct preferences and evaluation criteria  (see Appendix \ref{app:case_studies}).
These cases expose critical divergences in how models judge quality across scientific, creative, educational, and technical domains: (1) \textit{Length Bias}: Some models prefer short answers over detailed ones, even when a deeper explanation is needed (Cases 1, 4, 5);
(2) \textit{Nuanced Quality Insensitivity}: Some models fail to discern high quality in literary expression, complex reasoning, or specialized technical material (Cases 1, 2, 5);
(3) \textit{Context Insensitivity}: Some models use the same standard to judge all responses, even for different kinds of tasks that might need detailed analysis, creative expression, or technical expertise (Cases 1, 2, 3, 5);

These systematic evaluation patterns reveal fundamental differences in how reward models assess content quality, highlighting the importance of careful model selection for domain-specific applications and the need for more nuanced evaluation frameworks that can recognize diverse forms of excellence across diverse domains.

\subsection{Improving Reward Models via PMDC} \label{sec:improving}
Beyond evaluation, PMDC also can improve RMs through targeted fine-tuning. The maximum discrepancy samples identified by PMDC represent precisely those ambiguous or challenging cases where reward models exhibit substantial disagreement, making them ideal candidates for high-leverage fine-tuning. To validate this hypothesis, we train a reward model (baseline) based on ArmoRM-Llama3-8B-v0.1 \citep{ArmoRM} with the same implementation strategy, and fine-tune it using the $4500$ oracle-annotated preference pairs selected by PMDC.

As shown in Table~\ref{tab:reward_improvement}, fine-tuning on PMDC-selected samples yields an overall performance gain of $3.1\%$ over the baseline on RewardBench2. The improvement is particularly notable in areas requiring nuanced judgment, such as \textit{Math} and handling of \textit{Ties}. Minor reductions are observed in \textit{Focus} and \textit{Safety}, likely due to the limited size and domain coverage of the fine-tuning set. Nevertheless, the overall improvement demonstrates that discrepancy-driven data selection can effectively boost reward model robustness and alignment fidelity. This result further validates the utility of PMDC not only as an evaluator but also as a data curation engine for reward modeling.

% \subsection{Revealing Model Biases via Case Studies} 
% Detailed qualitative case studies examining evaluator biases (\eg, conciseness versus contextual depth, technical specificity, \etc) are provided in Appendix~\ref{app:case_studies} due to space constraints.
\begin{table}[t]
\centering
\small
\renewcommand{\arraystretch}{1.1}
\setlength{\tabcolsep}{6pt}
\begin{tabular}{lccc}
\toprule
\textbf{Dimension} & \textbf{Baseline} & \textbf{Fine-tuned} & \textbf{Gain} \\
\midrule
Factuality & 0.534 & 0.545 & +1.1\% \\
Focus & 0.642 & 0.588 & -5.4\% \\
Math & 0.516 & 0.561 & +4.5\% \\
Precise IF & 0.366 & 0.381 & +1.5\% \\
Safety & 0.780 & 0.764 & -1.6\% \\
Ties & 0.570 & 0.641 & +7.1\% \\
\midrule
\textbf{Overall} & \textbf{0.568} & \textbf{0.599} & \textbf{+3.1\%} \\
\bottomrule
\end{tabular}
\caption{Reward model performance before and after fine-tuning on PMDC-identified samples.}
\label{tab:reward_improvement}
\end{table}

\section{Conclusion and Future Work}
This work introduces the PMDC, a dynamic and efficient framework for evaluating RM generalization, which is often neglected in conventional static benchmarks.
By adaptively selecting high-discrepancy response pairs from a diverse and open-domain prompt pool and employing an LLM-based oracle for scalable preference judgment, PMDC enables robust evaluation of RM generalization.
Empirical evaluation of $10$ RMs not only reveals significant ranking inconsistencies compared to traditional benchmarks but also uncovers nuanced model-specific strengths and weaknesses. These results affirm PMDC’s capacity to provide more cost-effective and behaviorally insightful assessments of reward models.

Future research could focus on enhancing oracle reliability through multi-judge ensembles and extending PMDC to multi-dimensional evaluation frameworks to better capture capability trade-offs. Scaling the framework to larger model sets and prompt pools would further enhance its robustness and applicability.

\section*{Limitations}

While PMDC offers a cost-effective framework for evaluating RM generalization, several limitations remain.
First, unlike conventional benchmarks that evaluate $N$ models independently with $O(N)$ complexity, PMDC relies on pairwise comparisons, resulting in $O(N^2)$ model pairs and additional scoring overhead, which may limit scalability for very large RM collections.
Second, PMDC depends on an LLM-based oracle whose inherent biases (e.g., stylistic preferences) may affect evaluation fidelity, though our experiments show high inter-judge agreement.
Third, despite efforts to diversify the prompt pool, coverage of highly specialized or rapidly evolving domains remains limited.

% \section*{ETHICS STATEMENT}
% This work focuses on evaluating reward models for aligning language models with human preferences. It does not involve the collection or analysis of personally identifiable, sensitive, or harmful data. All datasets used are publicly available benchmarks or synthetically generated using licensed models, with no private user content included. Human-like judgments were simulated via LLM-based oracles under carefully designed prompts to minimize bias and ensure consistency; no real human annotators were involved in this study. The proposed PMDC framework is intended solely for academic research and responsible model evaluation, with no foreseeable misuse toward generating, promoting, or optimizing harmful, deceptive, or unethical content.

% \section*{Reproducibility Statement}
% We are committed to ensuring the reproducibility of our work. Upon publication, we will publicly release the complete source code, and detailed configuration files for all experiments. Additionally, we provide the prompt template for the LLM-based oracle in the appendix. These resources will enable the research community to reproduce our results, validate our findings, and extend the PMDC framework for future reward model evaluation and development.

\bibliography{custom}

\clearpage
\appendix

\section*{Appendix}
\setcounter{table}{0}   
\setcounter{figure}{0}
\setcounter{section}{0}
\setcounter{equation}{0}
\renewcommand{\thetable}{A\arabic{table}}
\renewcommand{\thefigure}{A\arabic{figure}}
\renewcommand{\thesection}{\Alph{section}}  
\renewcommand{\theequation}{A\arabic{equation}}

\section{Oracle Prompt}
\label{app:oracle_prompt}

\begin{tcolorbox}[colback=blue!3, colframe=black!70, arc=2mm, left=3pt,right=3pt,
    boxsep=5pt, boxrule=0.5pt, fonttitle=\bfseries, title=Prompt for Oracle Judge Evaluation]
\small
\textbf{System Message:} You are a professional text quality assessment expert. Please carefully compare the quality of two answers, focusing on: 1) Accuracy -- whether the information is correct; 2) Relevance -- whether it addresses the question; 3) Clarity -- whether the expression is clear and understandable; 4) Conciseness -- whether it is concise and avoids redundancy; 5) Depth -- whether it has insights; 6) Logic -- whether it is well-organized; 7) Practicality -- whether it is helpful to the questioner. Find the best balance between information content and readability. Only return the result in JSON format, without any explanation.

\textbf{User Prompt:} Please judge which of the following two responses is better. Only return the result in JSON format without any explanation.

Question: \{question\}

Response 1: \{response1\}

Response 2: \{response2\}

Please answer strictly in the following JSON format:
\{"preference": 1\} or \{"preference": 2\}

Where 1 means Response 1 is better, and 2 means Response 2 is better.
\end{tcolorbox}

\section{Random Sampling Results}
\begin{figure}[h]
    \centering
    \includegraphics[width=0.45\textwidth]{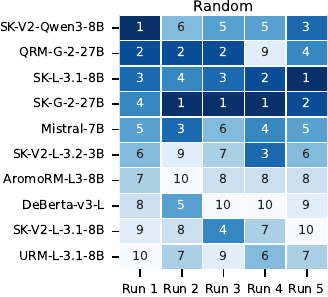}
    \caption{Random sampling's rank across 5 independent runs. The heatmap shows the rank of each RM in each run, with rank values annotated in individual cells.}
    \label{fig:random_consistency}
\end{figure}

\section{Use of Large Language Models}
Large language models were used as assistive tools during the preparation of this manuscript. Specifically, they helped refine language, improve phrasing, and enhance overall readability. All LLM-generated suggestions were carefully reviewed, verified, and edited by the authors.

\section{Implementation of PMDC} \label{app:alg}
\begin{algorithm}[H]
\caption{Pairwise Maximum Discrepancy Competition (PMDC)}
\label{alg:pmdc}
\small
\begin{algorithmic}[1]
\State \textbf{Input:} A Prompt pool, An LLM pool, Reward models $\mathcal{R} = \{R_1, \ldots, R_n\}$,  Oracle $\mathcal{O}$, Top-$k$ parameter $k$.
\State \textbf{Initialize:} Pairwise win counts $W_{ij} = 0$ for all $i,j \in \{1, \ldots, n\}$.

\State Sample a batch of prompts $\mathcal{Q}$ from the Prompt pool
\State Sample a batch of response generators $\mathcal{G}$ from the LLM pool

\For{each prompt $q$ in $\mathcal{Q}$}
    \State Generate response set $\{a_1, \ldots, a_m\}$ using generators $\mathcal{G}$.
    \State Compute all reward scores $s_{i}(q, a_j)$ for all $R_i \in \mathcal{R}$ and all $a_j$.
    \State Normalize scores: $s'_{i}(q, a_j) = \text{Min-Max}(s_{i}(q, a_j))$ for each model $R_i$.
    \State \textbf{Find Maximum Discrepancy samples:}
    \State Initialize discrepancy set $\text{MD}_{samples} = \emptyset$.
    \For{each model pair $(R_A, R_B) \in \binom{\mathcal{R}}{2}$}
        \State Compute discrepancies $D_{A,B}(q, a_i, a_j)$ for all response pairs $(a_i, a_j)$ using Eq.~\ref{eq:discrepancy}.
        \State $\text{MD}_{\text{pair}} \leftarrow \text{top-}k\{(D_{A,B}, q, a_i, a_j, R_A, R_B) : \forall i < j\}$.
        \State $\text{MD}_{samples} \leftarrow \text{MD}_{samples} \cup \text{MD}_{\text{pair}}$.
    \EndFor
    \State Total samples: $|\text{MD}_{samples}| = \binom{n}{2} \times k$.
    \State \textbf{Oracle Adjudication:}
    \For{each sample $(D, q, a_i, a_j, R_A, R_B)$ in $\text{MD}_{samples}$}
        \State Get oracle preference $P_{\mathcal{O}} = \mathcal{O}(q, a_i, a_j)$.
        \State Get model preferences $P_A = (s'_{A}(a_i) > s'_{A}(a_j))$ and $P_B = (s'_{B}(a_i) > s'_{B}(a_j))$.
        \If{$P_A = P_{\mathcal{O}}$ and $P_B \neq P_{\mathcal{O}}$}
            \State $W_{AB} \leftarrow W_{AB} + 1$.
        \ElsIf{$P_B = P_{\mathcal{O}}$ and $P_A \neq P_{\mathcal{O}}$}
            \State $W_{BA} \leftarrow W_{BA} + 1$.
        \EndIf
    \EndFor
\EndFor
\State \textbf{Global Ranking:}
\State Compute win-rate matrix $P$ from $W$ using Eq.~\ref{eq:winrate}.
\State Estimate BT ranking score $\boldsymbol{\xi}$ by maximizing Eq.~\ref{eq:bt_likelihood} using $W$.
\State \textbf{Output:} Global ranking based on $\boldsymbol{\xi}$.
\end{algorithmic}
\end{algorithm}

% No unverified content was included, and all scientific reasoning, experimental design, analysis, and conclusions were solely the work of the authors.

\section{Ethics Statement}
This work focuses on evaluating reward models. It does not involve the collection or analysis of personally identifiable, sensitive, or harmful data. All datasets used are publicly available benchmarks or synthetically generated using licensed models, with no private user content included. 
% The proposed PMDC framework is intended solely for academic research and responsible model evaluation, with no foreseeable misuse toward generating, promoting, or optimizing harmful, deceptive, or unethical content.

\section{Cases} \label{app:case_studies}
Below are five representative cases from our empirical evaluation data that reveal how different reward models exhibit distinct preferences and evaluation criteria.

\begin{table*}[!h]
\centering

\footnotesize
\begin{tabularx}{\textwidth}{XX}
\toprule
\multicolumn{2}{p{13.5cm}}{\textbf{Prompt:} \newline What will happen if one lemming jumps off a cliff?} \\ 
\midrule

\textbf{Response A (Skywork Preference - Shorter):\newline} 
If one lemming jumps off a cliff, the outcome would depend on the specific circumstances, but here's what typically happens:\newline
{\color{red}\textbf{1. In Reality}}: Contrary to the popular myth, lemmings do not commit mass suicide by jumping off cliffs. This misconception was popularized by the 1958 Disney documentary *White Wilderness*, which staged the scene for dramatic effect.\newline
{\color{red}\textbf{Migration \& Accidents}}: Lemmings may occasionally fall off cliffs or drown while migrating in large groups, especially when crossing unfamiliar terrain or bodies of water. However, this is accidental, not intentional.\newline
{\color{red}\textbf{2. If a Single Lemming Falls}}: A lemming might survive a short drop if the cliff isn't too high and it lands safely (e.g., in snow or vegetation). However, a significant fall could injure or kill it.\newline
{\color{red}\textbf{No Group Effect}}: Unlike the myth suggests, one lemming jumping would not trigger others to follow. Lemmings are not blindly obedient creatures.\newline
{\color{red}\textbf{3. Ecological Impact}}: A single lemming's death would have minimal impact on the population, as they reproduce quickly. However, predators (like owls or foxes) might scavenge the carcass.\newline
\textbf{Key Takeaway}: The idea of lemmings mindlessly plunging to their deaths is a myth. In reality, they're simply small rodents that sometimes face hazards during migration—just like other animals.
& 
\textbf{Response B (Oracle Preference - Comprehensive):\newline}  
This is a fantastic question because it touches on one of the most persistent myths in the animal kingdom. The answer has two parts: the literal, physical reality and the debunking of the famous myth.\newline
{\color{green!80!black}\textbf{The Literal, Physical Answer}}: If one lemming jumps off a cliff, it will fall. What happens next depends entirely on physics: If it's a short drop onto soft ground or into deep water, the lemming might survive. They are decent swimmers. If it's a high drop onto rocks or hard ground, the lemming will almost certainly be injured or killed by the impact.\newline
Crucially, {\color{green!80!black}\textbf{nothing else will happen. The other lemmings will not mindlessly follow it over the edge}}.\newline
{\color{green!80!black}\textbf{The Myth vs. The Reality}}: The popular image of lemmings lining up to commit mass suicide by leaping off cliffs is completely false. This idea was created and popularized by a very specific source.\newline
{\color{green!80!black}\textbf{Where Did the Myth Come From?}}: The myth was cemented in the public consciousness by the 1958 Disney nature film, ``White Wilderness.'' The filmmakers brought lemmings to a cliff, used a spinning turntable to herd and force the frantic animals over the edge, and used clever editing to make it look like a willful, suicidal stampede. The scene was entirely staged.\newline
{\color{green!80!black}\textbf{What Really Happens in Nature?}}: Lemmings have dramatic population cycles. Every 3-4 years, their numbers explode, leading to mass migration to find new food and territory. During these migrations, accidental deaths occur when they try to cross obstacles like rivers and cliffs. These are accidental deaths, not intentional suicide.
\\ \midrule
\multicolumn{2}{p{13.5cm}}{\textbf{Oracle Judgment}: Response B (comprehensive myth-busting with detailed scientific explanation) \newline \textbf{Model Discrepancy}: Skywork-V2-Llama-3.1-8B (A: 0.907 vs B: 0.319) strongly favored the shorter response, while ArmoRM (A: 0.596 vs. B: 0.745) and Oracle preferred the detailed scientific analysis.}\\ \midrule
\multicolumn{2}{p{13.5cm}}{
\textbf{Response A's Limitations}: While factually correct, it provides only surface-level debunking without explaining the myth's origins or providing comprehensive scientific context. The explanation lacks depth in describing the actual biological behaviors of lemmings. \newline
\textbf{Response B's Advantages}: Offers a structured, educational approach with clear separation between physical reality and myth debunking. Provides detailed historical context about the Disney film's staged footage, explains the actual population cycle mechanisms, and distinguishes between accidental deaths and intentional behavior. The response demonstrates superior scientific rigor and educational value.}\\
\bottomrule
\end{tabularx}
\caption*{\textbf{Case 1}: Skywork-V2-Llama-3.1-8B vs. ArmoRM-Llama3-8B-v0.1 comparison on scientific myth-busting. The former exhibits bias favoring brevity over comprehensive scientific analysis.}
\label{tab:case-study-1}
\end{table*}
\clearpage

\begin{table*}[t]
\centering

\footnotesize
\begin{tabularx}{\textwidth}{XX}
\toprule
\multicolumn{2}{p{13.5cm}}{\textbf{Prompt:} \newline Write a funny, interesting, inspiring poem for Women's Day.} \\ 
\midrule

\textbf{Response A (Oracle Preference - Modern \& Bold):\newline} 
\textbf{``Ode to the Glorious Glitch in the System''}\newline
{\color{green!80!black}\textbf{Oh, Woman! You magnificent force of nature}},\newline
With your multitasking, your grace, and your *attitude*.\newline
You've changed the world with a laptop, a spatula, and a laser,\newline
While raising kids, running nations, and inventing the *attitude*.\newline
{\color{green!80!black}\textbf{You were told to sit, so you stood.}}\newline
{\color{green!80!black}\textbf{Told to whisper, so you *roared*.}}\newline
Told to stay in the kitchen—\newline
So you turned it into a startup *and* baked a cake *and* ignored.\newline
{\color{green!80!black}\textbf{You've coded rockets, cured diseases}},\newline
Led revolutions in heels (and sometimes barefoot).\newline
You've balanced budgets, babies, and boardroom decisions,\newline
All while being told, ``Smile more, don't be so *ruthless*—wait, no, be *shrewd*?''\newline
{\color{green!80!black}\textbf{You are not ``just'' a mother, a wife, or a ``lady.''}}\newline
{\color{green!80!black}\textbf{You are *also* a storm, a sonnet, a spark, and a scheme.}}\newline
You are Beyoncé on stage, Malala in class,\newline
And Aunt Linda who *still* won the neighborhood pickleball team.\newline
So here's to the women who *do*, who *dare*, who *demand*,\newline
{\color{green!80!black}\textbf{Keep being weird, wise, wild, and wonderful.}}\newline
The world doesn't stand a chance.\newline
*(And honestly? It never did.)* 
& 
\textbf{Response B (V2 Preference - Traditional \& Rhyming):\newline}  
Here's to the woman, a creature of myth,\newline
Whose handbag contains what the universe smiths.\newline
{\color{red}A charger, a lip balm, a singular sock,}\newline
{\color{red}A half-eaten biscuit as hard as a rock},\newline
A receipt from two-thousand-and-three,\newline
And the keys to a car that she sold. Who is she?\newline
{\color{red}She's a master of multitasking arts,}\newline
{\color{red}Juggling deadlines and feelings and hearts.}\newline
Her brain is a browser with sixty-four tabs,\newline
(Including the one tracking neighborhood cats).\newline
She carries the echo of those who once fought,\newline
The suffragette's spirit that couldn't be bought.\newline
{\color{red}She can be a soft dawn or a hurricane's might,}\newline
{\color{red}A comforting whisper or blaze burning bright.}\newline
She builds up an empire and builds up a friend,\newline
On her wisdom and strength, you can always depend.\newline
So here's to her laughter, her power, her soul,\newline
The one who makes broken things perfectly whole.\newline
{\color{red}Happy Women's Day! Now go find your crown.}\newline
(It's probably under a cushion somewhere in the lounge.)
\\ \midrule
\multicolumn{2}{p{13.5cm}}{\textbf{Oracle Judgment}: Response A (modern, bold voice with contemporary feminist themes) \newline \textbf{Model Discrepancy}: Skywork-Reward-Llama-3.1-8B (A: 0.288 vs. B: 0.096) favored Response A, while Skywork-Reward-V2-Llama-3.1-8B (A: 0.275 vs. B: 0.703) favored Response B, and Oracle preferred the bold creative approach.}\\ \midrule
\multicolumn{2}{p{13.5cm}}{
\textbf{Response A's Advantages}: Feels more inspiring and empowering while maintaining humor. It takes more creative risks with its structure and language, creating a piece that's both entertaining and genuinely moving. The range of references (from Beyoncé to Malala to ``Aunt Linda'') makes it more inclusive and celebratory of all types of women. Demonstrates bold creativity and emotional impact. \newline\textbf{Response B's Limitations}: Beautifully crafted but plays it safer. While charming with traditional rhyme schemes and consistent meter, it doesn't reach the same emotional heights or creative boldness as Response A. The divergent model preferences reveal fundamentally different aesthetic philosophies in creative content evaluation.}\\
\bottomrule
\end{tabularx}
\caption*{\textbf{Case 2}: Skywork-Reward-Llama-3.1-8B vs. Skywork-Reward-V2-Llama-3.1-8B comparison on creative poetry evaluation. The latter exhibits bias in aesthetic preferences affecting creative content evaluation.}
\label{tab:case-study-2}
\end{table*}

\clearpage

\begin{table*}[t]
\centering

\footnotesize
\begin{tabularx}{\textwidth}{XX}
\toprule
\multicolumn{2}{p{13.5cm}}{\textbf{Prompt:} \newline Context: [header] How to stop an aggressive cat [title] Make a loud noise to distract the cat. [step] A noise can distract the cat and help redirect it away from its aggressive behavior. You could try clapping your hands, banging two objects, or ringing a bell.\newline\newline Which ending makes the most sense?\newline A. [substeps] Making loud sounds to try to distract the cat can also distract it from other, more encouraging behavior. If the cat becomes aggressive, it may become defensive.\newline B. Direct the cat towards the noise to stop it and help it calm down. [substeps] Common loud noises to attract attention include a roaring vehicle or engine.\newline C. You could also use your voice, which might get the cat to come near you so that it more easily hears you. [substeps] Give the cat a low-pitched " ha-ha " sound.\newline D. [substeps] Cats can also be fearful of noises, so only use a loud noise if the cat is acting aggressively and you're afraid it could hurt someone, including another animal. If you use a loud noise to distract the cat, make sure it has a clear escape route in case it wants to flee.} \\ 
\midrule

\textbf{Response A (QRM-Gemma Preference - Comprehensive Analysis):\newline} 
The most logical and complete ending is **D**. Here's why:\newline
{\color{green!80!black}\textbf{- D provides important context}} about when and how to use loud noises to stop an aggressive cat. It warns that cats can be fearful of loud sounds and emphasizes that this method should only be used when necessary (if the cat is a danger to others). It also advises ensuring the cat has an escape route, which is crucial for preventing further stress or aggression.\newline
{\color{green!80!black}\textbf{- A is contradictory}} — it suggests loud noises might distract from ``encouraging behavior'' but doesn't clarify what that means, and it implies the cat may become defensive, which isn't helpful advice for stopping aggression.\newline
{\color{green!80!black}\textbf{- B is illogical}} — directing a cat *toward* a loud noise would likely increase stress, not calm it down. The example of a ``roaring vehicle or engine'' is also impractical and overly extreme.\newline
{\color{green!80!black}\textbf{- C shifts focus to using the owner's voice}}, which doesn't align with the original step (using loud noises to distract). The suggested ``ha-ha'' sound is unclear and not a standard cat behavior modification technique.\newline
{\color{green!80!black}\textbf{Thus, D is the best choice}} because it offers practical, safety-conscious advice while staying relevant to the original step.
& 
\textbf{Response B (Skywork Preference - Similar but with Style Elements):\newline}  
The most logical and contextually appropriate ending is:\newline
{\color{red}\textbf{D. [substeps] Cats can also be fearful of noises...}}\newline
{\color{red}\textbf{Reasoning:}}\newline
{\color{red}- The original step recommends using a loud noise} (like clapping or ringing a bell) to **distract** an aggressive cat.\newline
{\color{red}- Option D provides a thoughtful, responsible follow-up}: it acknowledges the potential fear response in cats, advises caution, and emphasizes safety for both people and the cat.\newline
{\color{red}- It aligns well with humane behavior management} by suggesting the noise be used only when necessary and ensuring the cat can retreat, reducing further stress or escalation.\newline
The other options are less appropriate:\newline
{\color{red}- **A** is confusing and contradictory} — distracting from ``encouraging behavior'' doesn't fit the context.\newline
{\color{red}- **B** suggests directing the cat *toward* the noise}, which is counterintuitive and potentially dangerous; also, a roaring engine is impractical and overly intense.\newline
{\color{red}- **C** introduces a strange ``ha-ha'' sound} that doesn't align with standard cat behavior advice and shifts focus oddly to attracting the cat.\newline
{\color{red}Thus, **D** is the best choice.} 
\\ \midrule
\multicolumn{2}{p{13.5cm}}{\textbf{Oracle Judgment}: Response B (clearer structure, better formatting, concise analysis) \newline \textbf{Model Discrepancy}: QRM-Gemma-2-27B (A: 0.817 vs. B: 0.883) favored Response A, while Skywork-Reward-V2-Llama-3.1-8B (A: 0.760 vs. B: 0.227) strongly favored Response A.}\\ \midrule
\multicolumn{2}{p{13.5cm}}{
\textbf{Response A's Limitations}: While providing accurate systematic analysis and good breakdown of each option, it lacks the organizational clarity and precise language of Response B. The presentation is more dense and harder to scan, despite containing solid analytical content. The divergent model preferences reveal Skywork's systematic bias against well-formatted educational content, failing to recognize how superior presentation enhances learning effectiveness.\newline \textbf{Response B's Advantages}: Demonstrates superior organization with clear formatting, bold headers, and structured presentation that enhances readability. Uses more precise language (``contextually appropriate'', ``thoughtful, responsible follow-up'') and explicitly mentions ``humane behavior management'' showing deeper understanding. The checkmark emoji and clean separation of reasoning sections make it easier to follow while maintaining analytical thoroughness. }\\
\bottomrule
\end{tabularx}
\caption*{\textbf{Case 3}: QRM-Gemma-2-27B vs. Skywork-Reward-V2-Llama-3.1-8B comparison on educational guidance assessment. The latter exhibits bias against comprehensive analytical responses.}
\label{tab:case-study-3}
\end{table*}
\clearpage

\begin{table*}[t]
\centering

\footnotesize
\begin{tabularx}{\textwidth}{XX}
\toprule
\multicolumn{2}{p{13.5cm}}{\textbf{Prompt:} \newline Describe a scenario where artificial intelligence could be used to improve the quality and efficiency of healthcare delivery.} \\ 
\midrule

\textbf{Response A (Gemma Preference - Focused Narrative):\newline} 
Certainly! Here's a detailed scenario illustrating how artificial intelligence (AI) can enhance healthcare quality and efficiency:\newline
{\color{red}\textbf{Scenario: AI-Powered Personalized Patient Management in a Hospital Setting}}\newline
{\color{red}\textbf{Context:}} A large urban hospital aims to improve patient outcomes and optimize resource allocation in its cardiology department, which handles a high volume of patients with varying degrees of heart disease severity.\newline
{\color{red}\textbf{How AI is Used:}}\newline
{\color{red}\textbf{1. Early Risk Detection and Diagnosis:}} Upon patient admission, AI algorithms analyze electronic health records (EHR), including medical history, lab results, imaging scans (like echocardiograms), and even wearable device data. The AI model identifies patients at high risk of adverse cardiac events by spotting subtle patterns that might be missed by clinicians.\newline
{\color{red}\textbf{2. Personalized Treatment Planning:}} Based on the risk assessment, the system suggests personalized treatment plans. For instance, it recommends medication adjustments, lifestyle interventions, or scheduling for specific diagnostic tests tailored to the patient's unique profile.\newline
{\color{red}\textbf{3. Real-Time Monitoring and Alerts:}} Wearable devices monitor patients' vital signs continuously. AI detects anomalies such as arrhythmias or sudden changes in blood pressure and immediately alerts healthcare providers.\newline
{\color{red}\textbf{Outcomes:}} Improved Patient Outcomes, Increased Efficiency, Cost Savings, Enhanced Patient Experience.
& 
\textbf{Response B (Oracle \& V2-Qwen Preference - Comprehensive Structure):\newline}  
Artificial intelligence (AI) has the potential to significantly enhance healthcare delivery by improving quality, efficiency, and accessibility. Here's a detailed scenario illustrating how AI could be applied:\newline
{\color{green!80!black}\textbf{Scenario: AI-Powered Early Disease Detection and Personalized Treatment in Primary Care}}\newline
{\color{green!80!black}\textbf{1. Patient Data Integration and Analysis}} - A patient visits their primary care physician with vague symptoms like fatigue, weight loss, and occasional dizziness. The clinic uses an **AI-driven electronic health record (EHR) system** that aggregates data from past visits, lab results, wearable devices, and genetic information.\newline
{\color{green!80!black}\textbf{2. Early Diagnosis and Risk Prediction}} - The AI detects subtle irregularities and flags a **high risk for Type 2 diabetes and cardiovascular disease**, even though traditional diagnostic thresholds haven't yet been crossed.\newline
{\color{green!80!black}\textbf{3. Personalized Treatment Planning}} - The AI suggests **personalized lifestyle modifications** and **precision medicine interventions** based on the patient's genetic profile.\newline
{\color{green!80!black}\textbf{4. Continuous Monitoring and Adaptive Care}} - Real-time data feeds enable **automatic scheduling** and **prediction of potential complications**.\newline
{\color{green!80!black}\textbf{5. Streamlining Administrative Tasks}} - AI automates **appointment scheduling, insurance pre-authorizations, and billing**.\newline
{\color{green!80!black}\textbf{Benefits:}} Early Detection, Reduced Costs, Personalized Care, Efficiency, Scalability.\newline
{\color{green!80!black}\textbf{Challenges:}} Data privacy, algorithmic bias, human oversight requirements.
\\ \midrule
\multicolumn{2}{p{13.5cm}}{\textbf{Oracle Judgment}: Response B (comprehensive structure, broader scope, practical considerations) \newline \textbf{Model Discrepancy}: Skywork-Reward-Gemma-2-27B (A: 0.622 vs. B: 0.912) strongly favored Response B, while Skywork-Reward-V2-Qwen3-8B (A: 0.923 vs. B: 0.383) strongly favored Response A with massive preference score discrepancy.}\\ \midrule
\multicolumn{2}{p{13.5cm}}{
\textbf{Response A’s Limitations}: While providing coherent cardiology-focused content, it demonstrates significant shortcomings in scope and depth. Offers generic references to ``subtle patterns'' and ``anomalies'' without concrete specificity. Lacks consideration of implementation challenges, administrative complexities, or ethical considerations. The narrow departmental focus misses broader healthcare system realities and presents an overly idealistic view without acknowledging practical barriers.\newline \textbf{Response B's Advantages}: Demonstrates superior content depth with comprehensive 5-stage coverage from data integration to administrative efficiency. Provides concrete specificity (actual symptoms, biomarkers, technologies) and excellent hierarchical organization with clear headers. Shows sophisticated understanding of healthcare complexities including privacy concerns, algorithmic bias, and human oversight requirements. Offers balanced perspective addressing both benefits and implementation challenges, making it more valuable for practical healthcare AI deployment. }\\
\bottomrule
\end{tabularx}
\caption*{\textbf{Case 4}: Skywork-Reward-Gemma-2-27B vs. Skywork-Reward-V2-Qwen3-8B comparison on healthcare AI applications. The latter exhibits bias against comprehensive technical content and practical implementation considerations.}
\label{tab:case-study-4}
\end{table*}
\clearpage

\begin{table*}[t]
\centering

\footnotesize
\begin{tabularx}{\textwidth}{XX}
\toprule
\multicolumn{2}{p{13.5cm}}{\textbf{Prompt:} \newline What are the wavelengths of light emitted by metals depending on their temperature?} \\ 
\midrule

\textbf{Response A (Skywork Preference - Traditional Academic):\newline} 
The wavelengths (or spectrum) of light emitted by metals depending on their temperature are primarily described by **blackbody radiation** principles. When a metal is heated, it emits electromagnetic radiation across a range of wavelengths, and the distribution of this emitted radiation depends mainly on its temperature.\newline
{\color{red}\textbf{Key Concepts:}}\newline
{\color{red}\textbf{1. Blackbody Radiation}} - Metals, when heated enough, approximate a *blackbody radiator*—an idealized physical body that absorbs all incident electromagnetic radiation and re-radiates energy characteristic of its temperature.\newline
{\color{red}\textbf{2. Planck's Law}} - The spectral radiance B($\lambda$,T) of a blackbody at temperature T (in kelvins) as a function of wavelength $\lambda$ is given by Planck's radiation law: [Complex mathematical formula with h, c, k\_B constants]\newline
{\color{red}\textbf{3. Wien's Displacement Law}} - The wavelength $\lambda$\_max at which the radiation is most intense (peak wavelength) is inversely proportional to the temperature: $\lambda$\_max = b/T where b $\approx$ 2.898 $\times$ 10$^{-3}$ m$\cdot$K is Wien's displacement constant.\newline
{\color{red}\textbf{Typical Wavelengths at Different Temperatures:}} At $\sim$500 K (227$^{\circ}$C): Peak emission in infrared (around 5.8 $\mu$m), invisible to human eye. Around 1000 K ($\sim$727$^{\circ}$C): Peak shifts to near-infrared ($\sim$2.9 $\mu$m), some thermal glow begins. At 1500 K ($\sim$1227$^{\circ}$C): Peak near 1.9 $\mu$m, metal starts to glow dull red. At 2000 K ($\sim$1727$^{\circ}$C): Peak around 1.45 $\mu$m, glowing brighter red-orange. At 3000 K ($\sim$2727$^{\circ}$C): Peak shifts near 1 $\mu$m, metal glows bright orange-white. At 5000 K ($\sim$4727$^{\circ}$C): Peak near 0.58 $\mu$m (yellow-green), very bright white light.\newline
{\color{red}\textbf{Practical Notes:}} Real metals do not emit as perfect blackbodies. Their emissivity (how efficiently they emit radiation) varies with wavelength and material. Thermal radiation spectrum is continuous, unlike atomic emission lines.
& 
\textbf{Response B (Oracle \& ArmoRM Preference - Interactive Educational):\newline}  
Of course! This is an excellent question that gets to the heart of a fundamental concept in physics called **black-body radiation**.\newline
The short answer is: the light emitted by a metal is a **continuous spectrum of wavelengths**, and the **peak wavelength** depends almost entirely on its **temperature**, not the specific type of metal.\newline
{\color{green!80!black}\textbf{The Key Principles}} - Any object with a temperature above absolute zero emits thermal radiation...\newline
{\color{green!80!black}\textbf{1. Wien's Displacement Law: The Color of the Glow}} - **The Formula:** $\lambda$\_peak = b/T. **What it means:** As temperature **increases**, the peak wavelength becomes **shorter**. This explains the color change: **Shorter Wavelengths** = Blue/Violet/UV, **Longer Wavelengths** = Red/Infrared.\newline
{\color{green!80!black}\textbf{2. Stefan-Boltzmann Law: The Brightness of the Glow}} - **The Formula:** P = $\varepsilon$ $\cdot$ $\sigma$ $\cdot$ A $\cdot$ T$^{4}$. **What it means:** As temperature increases, brightness increases dramatically.\newline
{\color{green!80!black}\textbf{Step-by-Step: Wavelength and Color vs. Temperature}} - Let's walk through what happens as you heat a piece of iron:\newline
{\color{green!80!black}\textbf{Comprehensive Temperature Table:}} **$\sim$25$^{\circ}$C**: **No visible light.** **$\sim$530$^{\circ}$C**: **Faint, Dull Red Glow.** **$\sim$1000$^{\circ}$C**: **Bright Cherry Red to Orange.** ... **$\sim$2500$^{\circ}$C**: **"White Hot"** (tungsten filament temperature). **$\sim$5500$^{\circ}$C**: **Bluish-White** (like the Sun).\newline
{\color{green!80!black}\textbf{Visualizing the Spectrum}} - [Includes blackbody radiation curve image] The **peak** shifts to shorter wavelengths as temperature rises, and **total brightness** increases massively.\newline
{\color{green!80!black}\textbf{Important Distinction}} - This continuous thermal radiation is **different** from atomic emission spectra. Metal incandescence produces a smooth, continuous spectrum determined by temperature.
\\ \midrule
\multicolumn{2}{p{13.5cm}}{\textbf{Oracle Judgment}: Response B (interactive educational approach, comprehensive table, visual aids) \newline \textbf{Model Discrepancy}: ArmoRM-Llama3-8B-v0.1 (0.021 vs. 0.694) strongly favored Response B, while Skywork-Reward-V2-Llama-3.2-3B (0.389 vs. 0.292) favored Response A with massive preference score discrepancy (0.770).}\\ \midrule
\multicolumn{2}{p{13.5cm}}{
\textbf{Response A's Limitations}: Adopts overly theoretical approach starting with complex mathematical formulas that intimidate non-specialists. Information organization is scattered with practical applications buried in theoretical discussions. Temperature examples lack systematic progression and memorable associations.\newline \textbf{Response B's Advantages}: Demonstrates superior pedagogical design with clear ``short answer'' to detailed exploration progression. Features comprehensive temperature-color table with vivid descriptions (``Faint, Dull Red Glow,'' ``White Hot'') and practical anchors (tungsten filament, solar surface). Uses ``What it means'' explanations that bridge theory to intuitive understanding. Includes visual learning aids and distinguishes thermal from atomic spectra. Skywork's preference for Response A reveals systematic failure to recognize that effective science education requires both mathematical rigor and pedagogical accessibility.}\\
\bottomrule
\end{tabularx}
\caption*{\textbf{Case 5}: ArmoRM-Llama3-8B-v0.1 vs. Skywork-Reward-V2-Llama-3.2-3B comparison on physics education content. The latter exhibits bias against comprehensive educational formatting and visual learning aids.}
\label{tab:case-study-5}
\end{table*}
\clearpage

\end{document}